\documentclass{article}
\usepackage[T1]{fontenc}
\usepackage{iclr2027_conference,times}

\usepackage{graphicx}
\usepackage{capt-of}
\usepackage{wrapfig}
\usepackage{needspace}
\usepackage{listings,booktabs,colortbl}
\usepackage{hyperref}
\usepackage{url}
\graphicspath{{figures/}}
\hypersetup{
  hidelinks,
  pdftitle={What Accuracy Hides in Learned Transformer Programs},
  pdfauthor={Jonathan Lee, Narmeen Oozeer, Philip Quirke},
  pdfsubject={Evaluation and analysis of learned Transformer Programs}
}

\definecolor{TPAccent}{HTML}{0E5257}
\definecolor{TPInk}{HTML}{172B30}
\definecolor{TPMuted}{HTML}{425B60}
\definecolor{TPBorder}{HTML}{C4D1D3}
\definecolor{TPSurface}{HTML}{F7F9F9}
\lstdefinestyle{programdisplay}{
  language=Python,
  basicstyle=\fontfamily{lmtt}\selectfont\color{TPInk},
  keywordstyle=\color{TPAccent},
  commentstyle=\color{TPMuted},
  stringstyle=\color{TPInk},
  emph={select,aggregate_sum},
  emphstyle=\color{TPAccent},
  columns=fullflexible,
  keepspaces=true,
  showstringspaces=false,
  upquote=true,
  breaklines=true,
  breakatwhitespace=false,
  frame=none,
  xleftmargin=0pt,
  xrightmargin=0pt,
  aboveskip=0pt,
  belowskip=0pt
}
\lstnewenvironment{programcode}[1][]{\lstset{style=programdisplay,#1}}{}
\newsavebox{\programdisplaybox}
\newenvironment{programbox}{%
  \par\addvspace{1pt}\noindent%
  \setlength{\fboxsep}{1.5pt}\setlength{\fboxrule}{0.4pt}%
  \begin{lrbox}{\programdisplaybox}%
  \begin{minipage}{\dimexpr\linewidth-2\fboxsep-2\fboxrule\relax}%
  \color{TPInk}%
}{%
  \end{minipage}\end{lrbox}%
  \fcolorbox{TPBorder}{TPSurface}{\usebox{\programdisplaybox}}%
  \par\addvspace{1pt}%
}
\newcommand{\programnote}[1]{%
  \par\vspace{1pt}{\color{TPBorder}\hrule height 0.4pt}%
  \vspace{1pt}\noindent{\normalfont\color{TPMuted}#1}\par%
}
\newsavebox{\programscoresbox}
\newlength{\resultstablewidth}
\newenvironment{programscores}[1][@{}lrr@{}]{%
  \par\centering\color{black}%
  \arrayrulecolor{black}%
  \setlength{\tabcolsep}{2pt}%
  \setlength{\aboverulesep}{1pt}\setlength{\belowrulesep}{1pt}%
  \begin{lrbox}{\programscoresbox}\begin{tabular}{#1}%
}{%
  \end{tabular}\end{lrbox}%
  \ifdim\wd\programscoresbox>\linewidth
    \resizebox{\linewidth}{!}{\usebox{\programscoresbox}}%
  \else
    \usebox{\programscoresbox}%
  \fi\par%
}

\title{What Accuracy Hides in\\Learned Transformer Programs}
\author{Jonathan Lee\\
\normalfont Martian
\And
Narmeen Oozeer\\
\normalfont Martian
\And
Philip Quirke\\
\normalfont Martian}
\iclrfinalcopy

\begin{document}
\maketitle
\fancyhead{}
\renewcommand{\headrulewidth}{0pt}

\begin{abstract}
Transformer Programs (TP)~\citep{fc23} constrain a Transformer's architecture so that its learned computation can be extracted as readable Python code.
TP expose learned computations, but token accuracy can hide incorrect complete answers.
In this work, we examine how program inspection provides insights beyond accuracy scores and helps guide training.
We first evaluated TP on nine algorithmic tasks.
We compared token accuracy with accuracy on complete output sequences and checked for failures missed by sampled tests.
Further, we assessed training data improvements informed by program inspection and failure analysis, finding gains in sequence accuracy on several tasks.
We examined a counting task where a program achieved 100\% accuracy on sampled test sequences but failed on another input using the symbols and lengths allowed during training.
The program counted correctly, but the required answer was missing from its output labels.
After adding training examples with larger counts, the retrained models passed exhaustive evaluation over the bounded domain.
In the \textsf{frequency ranking} task, inspection showed how correct counts contributed to incorrect answers in programs trained before and after changes to the training data.
Finally, we studied constraints imposed by TP training and operations.
The TP training process can lead to accuracy losses late in training.
Our reference constructions also show how composing the restricted TP operations can use many layers or large lookup tables.
\end{abstract}

\section{Introduction}
\label{sec:introduction}

\begin{wrapfigure}{R}{0.44\textwidth}
\centering
\includegraphics[width=\linewidth]{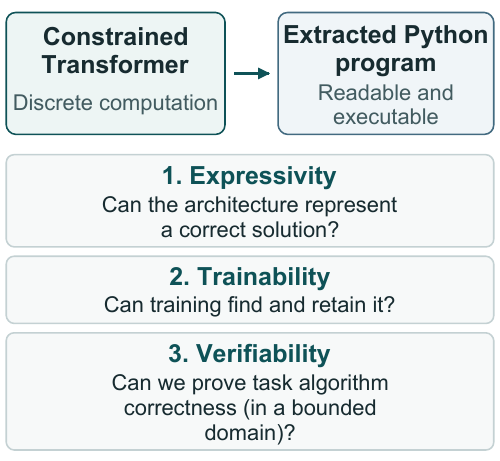}
\caption{We consider three distinct abilities: architectural expressivity, learning during training, and algorithmic verification.}
\label{fig:requirements}
\end{wrapfigure}
Our goal is to study programs where we can inspect the computations and whose correctness we can check over a defined input domain.
Transformer Programs (TP) constrain a Transformer's architecture so that its learned computation can be extracted as readable, executable Python code~\citep{fc23}.
This enables us to follow the calculations in the extracted code and compare its outputs with the required answers.
Having the code available allows us to understand the underlying computations and verify whether the program is correct giving interpretability by construction.

In order to establish correctness for a learned program, we require expressivity, trainability and verifiability (Figure~\ref{fig:requirements}).
First, the architecture should be expressive enough to represent a correct program for a given task.
However, just because a program can be represented does not mean that it will be learned.
The second requirement is trainability.
During training, soft choices are gradually sharpened toward the discrete choices required for program extraction.
Training must find a correct program and retain its performance as these choices become discrete.
The third requirement is verifiability, which allows us to establish whether the program achieves correctness over the claimed domain.

Existing work evaluates TP using token accuracy~\citep{fc23,blog}.
For the tasks we study, a complete answer receives a score of one only if every scored output token is correct, and zero otherwise.
We report this sequence accuracy as well as token accuracy because a partly correct output does not solve the task.
Several tasks achieve high token accuracy but much lower sequence accuracy.
Even perfect sequence accuracy on a sampled test set does not guarantee correctness.
One of our \textsf{histogram} programs, which labels each input value with the number of times it appears, achieves 100\% sequence accuracy on its sampled test set.
However, it fails on seven copies of 4, an input using the symbols and lengths allowed during training.
The program correctly counts seven occurrences, but seven is not among its available answer labels.
It instead returns six at every position.
A failure that the sampled test set missed is revealed by checking this input, while inspecting the computation allows us to discover why the answer is wrong.

We evaluate nine algorithmic tasks (Table~\ref{tab:benchmark-baselines}) and examine the \textsf{histogram} and \textsf{frequency ranking} programs in detail.
We trace their computations on failing inputs to explain how they produce incorrect answers.
For \textsf{histogram}, the diagnosis motivates adding training examples with larger counts.
For \textsf{frequency ranking}, we compare programs trained before and after changes motivated by earlier error analysis, examining which computations changed and which failures remain.
We use training data enrichment to emphasize rare or missing cases.
We make the following three contributions:

\begin{enumerate}
\item \textbf{Evaluation of TP with complete answers and a formal verifier.}
We evaluate token accuracy alongside accuracy on complete output sequences and use a formal verifier to search for violations of task specifications, checking counterexamples by program execution.
High token accuracy can conceal incorrect complete answers.
Counterexamples also reveal failures in programs that achieve perfect sequence accuracy on sampled tests.

\item \textbf{Program analysis and informed training improvements.}
We use computational diagnoses and observed failures to guide changes to training data.
For \textsf{histogram}, adding examples with larger counts produces retrained models that answer every input correctly within the checked domain.
Training changes that rebalance operand lengths also improve \textsf{addition} and \textsf{addition with hints}.
Our \textsf{frequency ranking} analysis explains how correct counts contribute to incorrect answers in programs trained both before and after enrichment.

\item \textbf{Analysis of constraints on the programs and their training.}
We show that accuracy can fall as training moves from soft choices toward the discrete choices required for executable code.
Our reference constructions also show how implementing some operations can require many layers or large lookup tables, but we do not claim that these constructions are minimal.
The cost of verification limits our correctness checks to small input domains.
Our \textsf{frequency ranking} analyses explain particular answers, but do not recover the full ranking algorithms.
\end{enumerate}

\section{Background}
\label{sec:background}

\subsection{Architecture}
\label{sec:architecture}

\begin{figure}[htbp]
\centering
\includegraphics[width=0.94\linewidth]{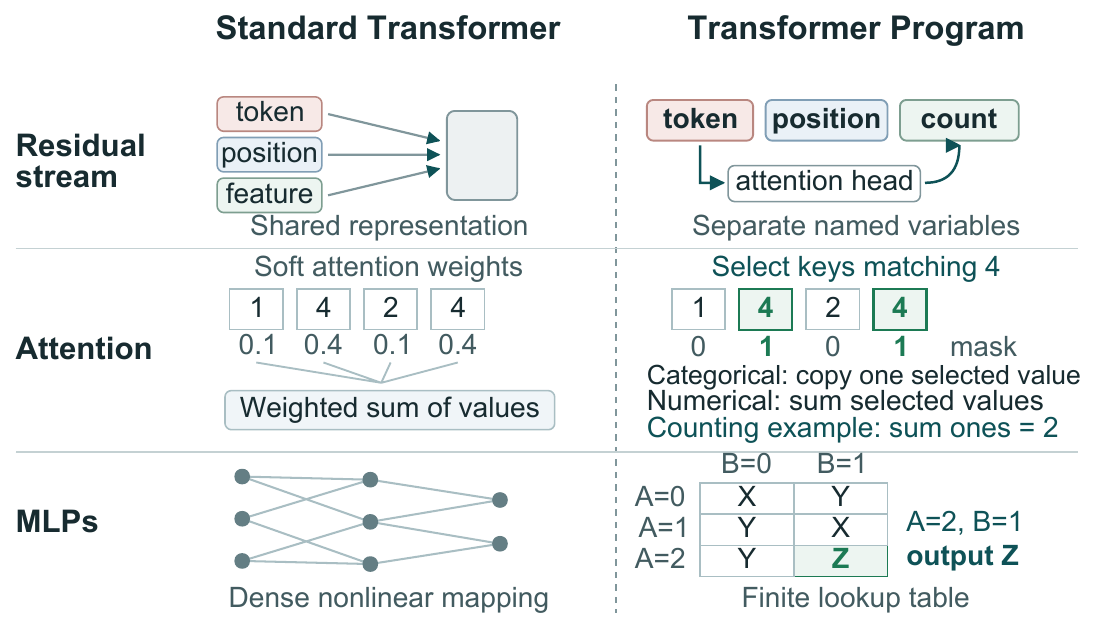}
\caption{By design Transformer Program models progressively restrict representations, attention and MLPs over training to finally learn discrete computations that enable code extraction.}
\label{fig:architecture-comparison}
\end{figure}
TP restrict how information is represented and how each component operates~\citep{fc23} (Figure~\ref{fig:architecture-comparison}).
The residual stream carries information between layers as separate named variables, which are either categorical or numerical.
A categorical variable can identify a token, while a numerical variable can store its count.
Categorical attention uses a learned matching rule to select a position and copy a value from it.
Numerical attention instead adds values from the positions that match the rule.
An MLP acts like a lookup table, taking variables from the same position and producing a category.
To choose the final answer, the classifier adds the contributions of the variables to each class score and selects the class with the largest total.
These constraints enable direct inspection of the learned computation as executable Python code.

During training, the model uses a continuous relaxation of its discrete choices allowing it to be optimized with gradient descent~\citep{fc23,gs17}.
At the start of training when the temperature is high, an attention head can use a mixture of candidate input variables rather than choose just one.
Reducing the temperature makes the choices sharper.
The hard neural model selects one alternative for each choice, while the classifier retains real valued weights.
Extraction converts the discrete computation into Python code.
We evaluate the soft model, the hard neural model and the extracted program separately because we cannot assume that all three give the same answers.

\subsection{Tasks and Reference Programs}
\label{sec:tasks}

Our experiments cover nine algorithmic tasks from the original TP benchmark and its reproduction study~\citep{fc23,blog}, and the program analysis is focused on \textsf{histogram} and \textsf{frequency ranking} tasks.
For \textsf{histogram}, the answer at each position is the number of occurrences of that input value.
The input \([1,5,1,2]\), for example, requires \([2,1,2,1]\), since 1 occurs twice and the other values occur once.
\textsf{Frequency ranking} returns each distinct value once ordered by decreasing count.
If two values have equal counts, the value appearing first in the input appears first in the output.
For \([0,1,1,2]\), the required answer is \([1,0,2]\).
We also write reference programs by hand to examine what the architecture can express.
A correct construction demonstrates that a solution exists for the specified input range, but does not show that training is able to find it.
The reference programs also do not show that the solution is the smallest possible solution.

\section{Better Benchmarking of TP}
\label{sec:benchmarking}

\subsection{Evaluation Protocol}
\label{sec:evaluation-protocol}

We provide sequence accuracy as well as token accuracy.
Sequence accuracy is determined by the fraction of examples in which every scored position is correct.
Scoring masks define which output positions count toward accuracy, including how padding and filler outputs are treated (Appendix~\ref{app:task-domains}).
Token accuracy for \textsf{frequency ranking}, \textsf{addition} and \textsf{addition with hints} was reported to be approximately 75\% to 92\%.
Sequence accuracy for these models was reported to be approximately 3\% to 40\% (Table~\ref{tab:benchmark-baselines}).

During training, TP learn using soft decisions, which become sharper as the temperature decreases.
These decisions are made discrete for hard evaluation and program extraction, but this does not guarantee that accuracy will be preserved.
For \textsf{addition}, an intermediate soft checkpoint achieved 90.50\% sequence accuracy.
The program extracted from that same checkpoint, however, achieved 14.16\% exact match on the answer digits.
We use the final checkpoint of an annealed run as our evaluation standard.

We cannot be certain that a program is correct just because it has perfect sequence accuracy on a test sample.
As a result, we use the existing Transformer Program Verification (TPV) framework~\citep{tpv} to search for errors within specified alphabet and input length bounds.
The framework encodes the checkpoint and task specification as logical constraints for the Z3 satisfiability modulo theories (SMT) solver.
We accept a counterexample only if running the extracted Python program on the example input confirms that its answer is wrong.
If the solver rules out constraint violations, we still need to establish that the encoding accurately represents the program over the checked domain.
Alternatively, we can evaluate the program on every input in the domain and check whether its outputs satisfy the task specification.
These assessments establish correctness only over the domain checked, not for longer inputs.
When we find an example that the program answers incorrectly, we inspect its computation to investigate why it gives the wrong answer.

\Needspace{23\baselineskip}
\subsection{Baseline Results}
\label{sec:baseline-results}

\begin{wraptable}{R}{0.53\textwidth}
\setlength{\belowcaptionskip}{\baselineskip}
\caption{Token and sequence accuracy for the selected best grid baselines, not averages across seeds. See Appendices~\ref{app:task-domains} and~\ref{app:training} for selection details and results across seeds.}
\label{tab:benchmark-baselines}
\centering
\footnotesize
\arrayrulecolor{black}
\setlength{\tabcolsep}{3pt}
\begin{tabular}{@{}lrr@{}}
\hline
Task & \shortstack{Token\\accuracy (\%)} & \shortstack{Sequence\\accuracy (\%)} \\
\hline
\textsf{Histogram} & 100.00 & 100.00 \\
\textsf{Sorting} & 100.00 & 100.00 \\
\textsf{Reversal} & 99.96 & 99.79 \\
\textsf{Dyck 1} & 99.89 & 98.60 \\
\textsf{Dyck 2} & 99.35 & 95.75 \\
\textsf{Double histogram} & 98.41 & 97.45 \\
\textsf{Frequency ranking} & 82.89 & 16.45 \\
\textsf{Addition} & 75.12 & 2.73 \\
\textsf{Addition with hints} & 91.52 & 39.66 \\
\hline
\end{tabular}
\end{wraptable}
Following the original study and reproduction study, our evaluation covered multiple model configurations and random seeds~\citep{fc23,blog}.
Our baseline grid covered nine algorithmic tasks, with 48 configurations and five seeds per configuration.
Each configuration combined one to four layers with 2, 4, 8 or 16 attention heads and 2, 4 or 8 MLPs per layer.
The head and MLP counts are totals, divided equally between categorical and numerical components.
We used the baseline training setup without the targeted interventions described in Section~\ref{sec:improving}.

We corrected two issues in the inherited \textsf{addition} generator before training.
Differences from prior reports~\citep{blog} therefore reflect changed examples as well as scoring conventions (Appendix~\ref{app:input-scoring}).
Table~\ref{tab:benchmark-baselines} gives both token and sequence accuracy for the selected grid baselines.
Baseline sequence accuracy for \textsf{double histogram} ranges from 32.0\% to 97.5\% across seeds.
\textsf{Addition with hints} and \textsf{frequency ranking} also vary across seeds (Appendix~\ref{app:training}).
Appendix~\ref{app:compute} summarizes compute use.

\subsection{Initial Audit}
\label{sec:initial-audit}

For \textsf{histogram}, every answer in the baseline's sampled test set was correct, but verification found an input on which the extracted program failed.
Given seven copies of 4, the task requires seven at each output position, but the program returns six throughout.
The input is within the training domain.
However, the perfect sampled score did not cover this failure (Appendix~\ref{app:verification}).
The next section examines what the learned programs compute.
Following their calculations on failing inputs helps us explain the errors and identify possible changes to training.

\section{Understanding the Learned Programs}
\label{sec:analysis}

We use \textsf{histogram} and \textsf{frequency ranking} as case studies to show how inspecting learned programs helps explain their failures.

\subsection{\textsf{Histogram} Before Enrichment}
\label{sec:hist-analysis}
\label{sec:hist-before}

For the input containing seven copies of four from Section~\ref{sec:benchmarking}, the program correctly counts seven at every ordinary position but returns six.
The classifier's available labels are one through six, which exclude the required answer (Figure~\ref{fig:histogram-missing-label}).
The following excerpt, extracted from the trained \textsf{histogram} model, correctly counts seven at every ordinary position:

\begin{programbox}
\begin{programcode}
# Relevant branch of num_predicate_0_0; other branches omitted.
elif q_token in {"4"}:
    return k_token == "4"

num_attn_0_0_pattern = select(tokens, tokens, num_predicate_0_0)
num_attn_0_0_outputs = aggregate_sum(num_attn_0_0_pattern, ones)
\end{programcode}
\programnote{For an input value of 4, \texttt{select} marks every position containing 4. \texttt{aggregate\_sum} adds one for each match, giving the number of occurrences of 4.}
\end{programbox}

\par\noindent\begin{minipage}[t]{0.53\linewidth}
\setlength{\parskip}{0.5pc}
\vspace{0pt}
The classifier weight file begins with this header: \lstinline[style=programdisplay,language={}]|feature,value,1,2,3,4,5,6|

The program takes its answer labels from those columns:
\begin{programbox}
\begin{programcode}
classes = classifier_weights.columns.to_numpy()
predictions = classes[logits.argmax(-1)]
\end{programcode}
\programnote{The first two header fields identify the feature and its value. The remaining fields name answer labels. Using the computed features, the classifier scores these labels. \texttt{argmax} selects the label with the highest score.}
\end{programbox}

\end{minipage}\hfill
\begin{minipage}[t]{0.44\linewidth}
\vspace{0pt}
\centering
\includegraphics[width=\linewidth]{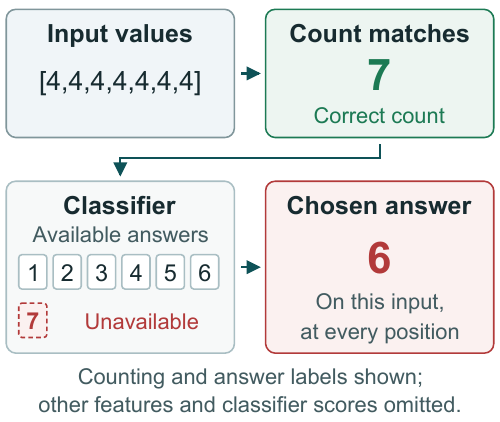}
\captionof{figure}{The program counts seven correctly but cannot output seven because that answer label is absent from its classifier.}
\label{fig:histogram-missing-label}
\end{minipage}\par

These labels come from the training data, which contained no examples requiring seven.
Including such examples would make seven available to the classifier.
Section~\ref{sec:improving} tests training changes motivated by this diagnosis.

\subsection{\textsf{Frequency Ranking} Before Enrichment}
\label{sec:most-freq-analysis}

Consider the input \([1,3,4,3,4,4,4]\).
The required answer is \([4,3,1]\): 4 appears four times, 3 appears twice and 1 appears once.
The answer should contain each distinct value once.
Instead, the baseline returns \([4,4,1]\), placing another 4 in the second position rather than 3.
To understand this error, we follow the calculation for that position.
The classifier scores each possible answer using contributions from the input and the intermediate values computed by the program.

In the operation below, both of the first two answer positions receive a count of four.
The count is correct because 4 occurs four times in the input.
It does not, by itself, specify which answer position should contain 4.

\begin{programbox}
\begin{programcode}
# Relevant branch of num_predicate_0_5.
# Position 0 is BOS. Other branches are omitted.
elif position in {1, 2}:
    return token == "4"

num_attn_0_5_pattern = select(tokens, positions, num_predicate_0_5)
num_attn_0_5_outputs = aggregate_sum(num_attn_0_5_pattern, ones)
\end{programcode}
\programnote{\texttt{select} marks the input positions containing 4. \texttt{aggregate\_sum} adds one for each match.}
\end{programbox}

We then examine how this count contributes to the scores for the second answer.

\begin{programbox}
\begin{programcode}
num_attn_0_5_output_scores = classifier_weights.loc[
    [("num_attn_0_5_outputs", "_") for v in num_attn_0_5_outputs]
].mul(num_attn_0_5_outputs, axis=0)
\end{programcode}
\programnote{The program retrieves this counting operation's learned weight for each possible answer and multiplies it by the count at each position.}
\end{programbox}

To see how this affects the choice, we separate the count contribution from all the other scores (Table~\ref{tab:baseline-frequency-scores}):

\begin{wraptable}{R}{0.56\linewidth}
\caption{Scores at the second output position in the baseline \textsf{frequency ranking} program, rounded to three decimal places. These are classifier scores, not counts or probabilities.}
\label{tab:baseline-frequency-scores}
\footnotesize
\begin{programscores}
\toprule
\textbf{Contribution} & \textbf{Answer 3} & \textbf{Answer 4} \\
\midrule
All other contributions combined & 15.894 & 15.797 \\
Contribution from the count of 4 & $-8.584$ & $-1.445$ \\
Final total & 7.310 & \textbf{14.353} \\
\bottomrule
\end{programscores}
\par
\normalsize
\setlength{\parskip}{0.5pc}
\end{wraptable}
The other contributions give the correct answer, 3, a slightly higher score than 4.
Adding the contribution from this count lowers the score for 3 more than it lowers the score for 4, reversing their order.
The final total for 4 is higher than for any other answer, so 4 appears a second time.
The error is therefore not in this counting operation, but in how its result contributes to the classifier's choice on this input (Appendix~\ref{app:most-freq-analysis}).
These score contributions are added separately for each output position.
The program then applies the label lookup and \texttt{argmax} procedure from Section~\ref{sec:hist-analysis} to return the answer with the largest total.
The following section describes the training changes and compares the resulting performance.

\section{Improving Performance}
\label{sec:improving}

\subsection{Improving Training}
\label{sec:intervention-method}

Before changing the training examples, we tested greater capacity and changes to training procedures.
Increasing depth to six layers did not improve \textsf{Dyck 1} and raised \textsf{Dyck 2}'s sequence accuracy from 95.75\% to 96.50\%, with failures remaining in both tasks.
Changes to training procedures~\citep{cho24,atp} had mixed results and did not consistently improve performance across seeds.
Appendix~\ref{app:training} reports the individual comparisons.
Increasing the amount of data was not sufficient either.
For \textsf{frequency ranking}, training on 50,000 naturally sampled examples achieved lower token accuracy compared to training on 20,000 examples.
We therefore explored whether changing the training examples could address the observed failures.
\textsf{Histogram} counterexamples contained larger counts that were rare or absent in training.
For \textsf{frequency ranking}, earlier error analysis found that failures often involved ties, motivating examples with tied counts.
The code analysis in Section~\ref{sec:analysis} was performed later; we use it here to compare the computations before and after the training changes.

In the case of \textsf{histogram}, examples containing counts of six and seven were mixed with randomly sampled inputs.
These were counts on which the baseline failed.
Including examples requiring seven also added seven to the classifier's available answer labels.
We retrained the entire model rather than updating only the classifier.

In the case of \textsf{frequency ranking}, examples with ties were created and added to training.
The mixture targeted 40\% of examples with three or more values sharing the highest count.
These values must be ordered by their first appearance in the input.
The training data was also greater in volume and the model was larger.
Further, the test set contained more ties.
Thus, the accuracy comparison involves several changes and does not isolate the effect of adding these training examples.

\subsection{Programs After Enrichment}

\subsubsection{\textsf{Histogram}}
\label{sec:hist-intervention}
\label{sec:hist-after}

For \textsf{histogram}, the verifier found counterexamples despite perfect sequence accuracy on the sampled test set.
To characterize these failures, we exhaustively evaluated a separate baseline model at the same configuration as the program examined in Section~\ref{sec:analysis}.
It answered 42 inputs incorrectly, all containing a value that appeared six or seven times.
Six was already an available answer label, so the missing label seven did not explain every failure.

\begin{figure}[htbp]
\centering
\includegraphics[width=\linewidth]{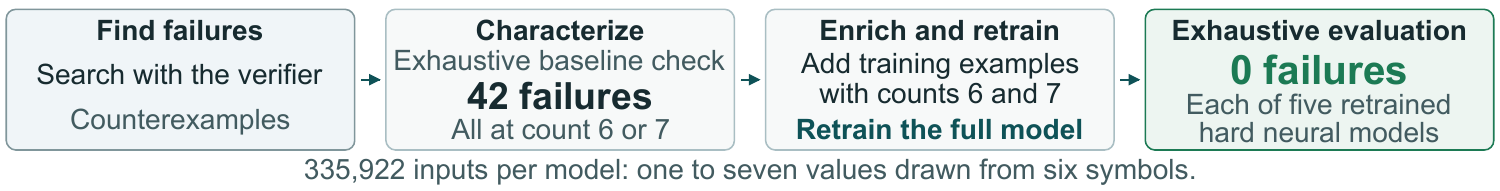}
\caption{\textsf{Histogram} counterexamples guide training data enrichment and full model retraining. All five retrained hard neural models answer every input correctly across the 335,922 inputs of one to seven values drawn from six symbols.}
\label{fig:histogram-repair-process}
\end{figure}

On the input containing seven copies of 4 from Section~\ref{sec:hist-analysis}, the retrained program now outputs seven at every ordinary position, producing the correct answer \([7,7,7,7,7,7,7]\) (Appendix~\ref{app:hist-program-records}).
After enriching the data with larger counts, we retrained five seeds at this configuration.
All five retrained hard neural models answered every input correctly across the 335,922 sequences of one to seven values drawn from six symbols (Figure~\ref{fig:histogram-repair-process}).

\subsubsection{\textsf{Frequency Ranking}}
\label{sec:most-freq-after}

We return to the input \([1,3,4,3,4,4,4]\) from Section~\ref{sec:most-freq-analysis}.
Its required answer is \([4,3,1]\).
The enriched program returns \([4,1,1]\).
Where the baseline repeats 4, the enriched program repeats 1.
To understand this change, we analyze how the enriched program derives its answer for the second position.
The enriched program contains a rule that counts occurrences of 3 at positions where the input value is 3.

\begin{programbox}
\begin{programcode}
# Relevant branch of num_predicate_0_5.
elif q_token in {"3"}:
    return k_token == "3"

num_attn_0_5_pattern = select(tokens, tokens, num_predicate_0_5)
num_attn_0_5_outputs = aggregate_sum(num_attn_0_5_pattern, ones)
\end{programcode}
\end{programbox}

\Needspace{20\baselineskip}
\begin{wraptable}{r}{0.56\textwidth}
\caption{Scores at the second output position in the enriched \textsf{frequency ranking} program, rounded to three decimal places. For answer 3, the weight is approximately $-4.064$, so the count subtracts 8.129 from its score. For answer 1, the weight is approximately 1.071, adding 2.142 instead.}
\label{tab:enriched-frequency-scores}
\footnotesize
\begin{programscores}
\toprule
\textbf{Contribution} & \textbf{Answer 3} & \textbf{Answer 1} \\
\midrule
All other contributions combined & 17.165 & 11.629 \\
Contribution from the count of 3 & $-8.129$ & 2.142 \\
Final total & 9.036 & \textbf{13.771} \\
\bottomrule
\end{programscores}
\par
\normalsize
\setlength{\parskip}{0.5pc}
\end{wraptable}
At the second output position, the input value is 3.
The rule therefore selects both occurrences of 3 and produces the correct count of two.
In the baseline program discussed in Section~\ref{sec:most-freq-analysis}, the operation shown supplies the count of 4 to this position instead.
The same weighting procedure discussed in Section~\ref{sec:most-freq-analysis} is employed by the enriched program, using its own learned classifier weights.
The other contributions favor answer 3 over 1 (Table~\ref{tab:enriched-frequency-scores}).
Adding the contribution from this count reverses their order.
With all contributions included, answer 1 has the highest score among the available answers and is returned at the second position.
The enriched program therefore still makes an error in which a correct count contributes to a wrong answer.

\Needspace{4\baselineskip}
\subsection{Final Results}
\label{sec:intervention-results}

\setlength{\resultstablewidth}{\textwidth}
\begin{table}[htbp]
\setlength{\belowcaptionskip}{\baselineskip}
\caption{Selected training results and verification outcomes. \(N\) includes special tokens. Clean verdicts apply only to the stated domains.}
\label{tab:final-results}
\centering
\footnotesize
\arrayrulecolor{black}
\setlength{\tabcolsep}{3pt}
\begin{tabular}{@{}>{\raggedright\arraybackslash}p{0.20\resultstablewidth}rrc>{\raggedright\arraybackslash}p{0.37\resultstablewidth}@{}}
\hline
Task & \shortstack{Token\\(\%)} & \shortstack{Seq.\\(\%)} & \shortstack{Counterexample\\found?} & Checked domain or first failure \\
\hline
\textsf{Histogram} & 100.00 & 100.00 & No & Exhaustive through \(N\leq8\) \\
\textsf{Sorting} & 100.00 & 100.00 & No & Exhaustive through \(N\leq8\) \\
\textsf{Reversal} & 99.96 & 99.79 & Yes & First failure at \(N=5\) \\
\textsf{Dyck 1} & 99.89 & 98.60 & Yes & Failure at trained length \(N=16\) \\
\textsf{Dyck 2} & 99.54 & 96.50 & Yes & Failure at trained length \(N=16\) \\
\textsf{Double histogram} & 100.00 & 100.00 & No & Checked through \(N\leq8\) \\
\textsf{Frequency ranking} & 86.18 & 57.60 & Yes & First failure at \(N=3\) \\
\textsf{Addition} & 94.09 & 76.30 & Yes & First failure at \(N=4\) \\
\textsf{Addition with hints} & 97.84 & 84.22 & Yes & Failure at \(N=8\) \\
\hline
\end{tabular}
\end{table}
\textsf{Addition} and \textsf{addition with hints} showed the largest reported improvements in sequence accuracy.
In the operand length experiments, \textsf{addition} improved from 2.73\% to 58.48\% and \textsf{addition with hints} from 39.66\% to 65.20\%.
The selected final recipes reached sequence accuracies of 76.30\% and 84.22\%, respectively.
A summary of the selected results from the training experiments is given in Table~\ref{tab:final-results}.
Token and sequence accuracy are based on sampled test sets.
The verification columns report whether an error was found and what domain was covered when no errors were found.
Only one of the five \textsf{sorting} seeds passed exhaustive evaluation under the selected training recipe (Appendix~\ref{app:repair}).
Some errors remained after the improvements.
\textsf{Addition} and \textsf{addition with hints} achieved better sequence accuracies, but verification still found incorrect answers.
\textsf{Double histogram} reached 100\% sequence accuracy for the selected program.
However, final sequence accuracy ranged from 32.0\% to 100.0\% across five seeds.
Improving the best result therefore did not mean that training was consistently successful.
Section~\ref{sec:constraints} examines constraints that help explain the remaining errors and instability in training.

\section{Constraints and Limitations}
\label{sec:constraints}

Our results illustrated how analyzing learned programs can explain a model's mistakes and prompt suggestions for changes to the training data.
We also observed constraints on learning accurate discrete programs and expressing computations with the available operations.
We cover these limits before discussing the scope of our correctness checks and computational explanations.

\Needspace{6\baselineskip}
\subsection{Training Toward Discrete Choices}

\Needspace{20\baselineskip}
\begin{wrapfigure}{R}{0.58\textwidth}
\centering
\includegraphics[width=\linewidth]{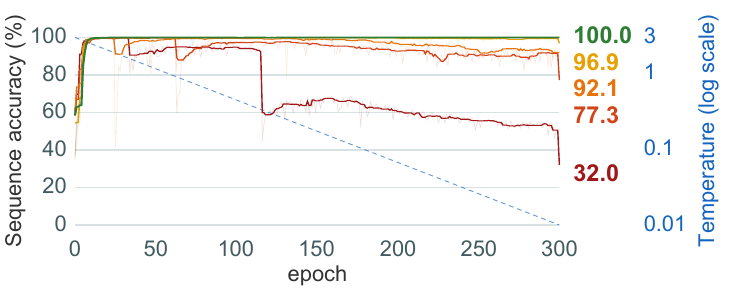}
\caption{\textsf{Double histogram} during annealing toward discrete TP choices. Four of five seeds lose accuracy from soft peaks above 99.89\%. Faint and solid curves show recorded and median filtered soft scores; the dashed blue curve shows temperature. Right labels give final program scores; only one retains 100\%.}
\label{fig:double-histogram-annealing}
\end{wrapfigure}
During training, the model may consider multiple possibilities and average them.
For instance, instead of selecting a single input variable, an attention head may use a weighted average of several candidate input variables.
The weights determine the contribution of each variable.
Lowering the temperature concentrates the weights on one singular variable.
This changes the information passed to later computations, which may no longer work as accurately as they did with the earlier mixture.

The \textsf{addition} example discussed in Section~\ref{sec:evaluation-protocol} illustrates both difficulties.
At epoch 71, the soft model achieved a sequence accuracy of 90.50\%.
However, the program extracted from this checkpoint achieved only 14.16\% exact match on the answer digits.
Selecting the checkpoint with the highest soft model sequence accuracy did not give an extracted program with comparable performance.
Also, continued training did not retain the peak performance.
The corresponding 500 epoch run ended at 56.12\% sequence accuracy, a decline of 34.38 percentage points from the soft peak.
These results show losses both when extracting discrete choices from a fixed checkpoint and when continuing training toward discrete choices.
The \textsf{double histogram} results illustrate variation among seeds.
After enrichment, the five seeds achieved soft sequence accuracies above 99.89\%.
However, four subsequently lost accuracy, and only one retained 100\% sequence accuracy in its final program (Figure~\ref{fig:double-histogram-annealing} and Appendix~\ref{app:annealing}).

\subsection{Costs of Reference Constructions}

The architecture places constraints on the behavior of components.
Attention selects and copies values or sums them.
MLPs choose output categories based on lookup tables.
Other computations are built by composing these components.
Our reference programs show that this can result in more layers or larger lookup tables.
Our \textsf{addition} construction passes carries between layers and uses each digit column to determine whether it generates, blocks or passes a carry.
If the digit column sum equals nine, its outgoing carry depends on the incoming carry from the previous column.
Resolving these dependencies uses \(n+2\) layers in our construction, where \(n\) is the maximum number of digits in an operand.
Ranking frequencies requires that values be counted and then ordered.
In the case of ties, the first occurring value gets precedence.
The equality test can be performed by the attention matching rule, but the rule cannot directly determine whether position \(i\) precedes position \(j\).
In our reference program, we use a lookup table for this computation.
The table takes the counts and first occurrence positions of the values and returns the required answer for the specified finite input range.
Its size grows roughly as \(N^V\), where \(N\) is sequence length and \(V\) is vocabulary size.
These constructions provide solutions to the tasks, but they may not be the most efficient methods (Appendix~\ref{app:constructions}).

\subsection{Scope of the Evidence}

Verification becomes more expensive with input length.
Our solver translates the model's operations and classifier weights into logical constraints to check learned programs.
For \textsf{double histogram}, checking inputs of total length seven took around 100 minutes.
At length eight, the solver did not finish within eight hours.
Thus, we directly executed the learned program on every input of that length instead.
The current encoding limits the input lengths we can check using the solver.
Our positive correctness results are restricted to the stated input ranges covered by the checks (Appendix~\ref{app:length-runtime}).

We studied the frequency programs before and after enrichment to understand their internal workings and how they arrive at their answers.
Both programs contain over 2,000 lines of Python and have 96 learned attention and MLP components each.
The snippets presented in this paper represent only a small fraction of these computations.
In the examples studied, both programs count correctly, but use those counts in a manner that favors an incorrect answer.
A count may contribute directly to the answer and also pass into later operations.
Lookup tables combine intermediate values, while different inputs may follow different branches.
Understanding the full program therefore requires more than reading each component separately.
Our analysis follows the computations for particular inputs and explains the resulting errors, but does not recover the full \textsf{frequency ranking} algorithms.
Our findings concern this architecture and the small algorithmic tasks studied here.

\section{Conclusion}
\label{sec:conclusion}

We evaluated TP by checking complete answers and analyzing the reasons for failures beyond the sampled test set.
High token scores often masked failures and even sampled sequence accuracy failed to identify some failures within the training domain.
Analyzing the programs helped explain these failures and showed how these computations contributed to the final answer.
Changes to training data informed by program inspection and counterexamples found from verification improved sequence accuracy on several tasks.
For the \textsf{histogram} task, we added training examples with larger counts.
The retrained models then provided correct answers for every input in the bounded domain checked by exhaustive evaluation.
These improvements do not change the distinction between expressivity, trainability and verifiability.
Training toward discrete choices can lead to poorer accuracy and our reference constructions used many layers or large lookup tables to implement certain computations.
Training results also differed across seeds.
Analyzing the frequency programs allowed us to identify the reasons for particular errors in the program, but did not recover the full ranking algorithms.
The correctness results are limited to the domains checked.
Beyond identifying incorrect answers, examining the programs helped us clarify specific reasons for failure, inform changes to training and check what those changes achieved.

\clearpage
\subsection*{AI Use Statement}
In this work, we used generative AI tools to assist with research implementation, experimental workflows, and the analysis and interpretation of results.
Additionally, we used generative AI tools to search for and summarize relevant literature, aid drafting and revising of the manuscript text, prepare scientific figures, and assist with LaTeX formatting and references.

The authors reviewed the manuscript, scientific claims, research code and experimental results.
We take responsibility for the final content of this work, including text, claims, code and other artifacts produced with the aid of generative AI.

\bibliographystyle{iclr2027_conference}
\bibliography{references}

\clearpage
\appendix
\section{Supporting Evidence and Methods}
\label{app:records}

\setcounter{table}{0}
\renewcommand{\thetable}{A\arabic{table}}
\renewcommand{\theHtable}{appendix.\arabic{table}}
\setcounter{figure}{0}
\renewcommand{\thefigure}{A\arabic{figure}}
\renewcommand{\theHfigure}{appendix.\arabic{figure}}

\subsection{Experimental Setup}
\label{app:task-domains}
The first seven tasks are from the Transformer Programs benchmark~\citep{fc23}.
The reproduction study by \citet{blog} adds addition and addition with hints.
Each of the nine tasks assigns an output to every input position.
Some positions in the output contain padding or filler, which is not part of the answer.
Answers to the tasks are defined in Table~\ref{tab:app-tasks}.
The required answers are defined independently of the labels provided in training examples.
For histogram, sampled training labels may stop at six; however, seven is still required when an input contains seven occurrences of a value.

\begin{table}[htbp]
\caption{Task specifications. Examples and definitions omit boundary tokens. The bracket tasks require a label for every prefix, not just a decision about the complete input.}
\label{tab:app-tasks}
\centering\footnotesize
\begin{tabular}{@{}>{\raggedright\arraybackslash}p{0.23\linewidth}>{\raggedright\arraybackslash}p{0.72\linewidth}@{}}
\toprule
Task & Required answer \\
\midrule
Histogram & At each position, the number of occurrences of its input value. \\
Double histogram & At each position, the number of distinct values with the same frequency as its input value. \\
Sorting & All input values in ascending order, preserving multiplicities. \\
Reversal & All input values in reverse order. \\
Frequency ranking & Each distinct value once, in decreasing frequency order. Equal frequencies are ordered by first occurrence. \\
Dyck 1 & At each prefix over one bracket type: balanced (\texttt{T}), valid but incomplete (\texttt{P}), or invalid (\texttt{F}). Invalidity persists through later prefixes. \\
Dyck 2 & The same prefix labels over two bracket types. A closing bracket must match the most recent unmatched opening bracket. \\
Addition & Decimal digits of the sum of two nonnegative operands, in their usual order. \\
Addition with hints & The same sum, with a letter before each operand and answer digit. Letters restart from \texttt{a} in each field and advance from left to right. \\
\bottomrule
\end{tabular}
\end{table}

Let \(U(x)\) be the distinct values in input \(x\), and let \(c_x(v)\) be the number of occurrences of value \(v\) in input \(x\).
At position \(i\), double histogram returns the number of values in \(U(x)\) that occur as often as \(x_i\).
Since double histogram considers distinct values, \([1,5,1,2]\) requires \([1,2,1,2]\).
Frequency ranking returns each distinct value once, in decreasing frequency order, with ties decided by first occurrence.
Its answer has length \(|U(x)|\).
Thus, \([5,4,5,4,1]\) requires \([5,4,1]\) since 5 occurs before 4.

\subsubsection{Input Domains and Scoring}
\label{app:input-scoring}
We let \(N\) denote the length of the input sequence before batch padding, including boundary tokens, delimiters and hint letters.
The beginning, end and padding tokens are \texttt{<s>}, \texttt{</s>} and \texttt{<pad>}, respectively.
Standard training input supports are defined in Table~\ref{tab:app-domains}.
The changes made for enrichment are described in Appendix~\ref{app:training}.
Appendix~\ref{app:verification} states the domains actually checked for correctness.

\begin{table}[htbp]
\caption{Standard input supports. Here \(m\) counts ordinary values or brackets, and \(d_a,d_b\) count operand digits. BOS and EOS denote beginning and end tokens.}
\label{tab:app-domains}
\centering\footnotesize
\begin{tabular}{@{}>{\raggedright\arraybackslash}p{0.25\linewidth}>{\raggedright\arraybackslash}p{0.44\linewidth}>{\raggedright\arraybackslash}p{0.22\linewidth}@{}}
\toprule
Tasks & Ordinary input and boundaries & Total length \\
\midrule
Sorting, reversal & Values 0 through 4; \(1\leq m\leq6\); BOS and EOS & \(N=m+2\), 3 to 8 \\
Histogram, double histogram, frequency ranking & Values 0 through 5; \(1\leq m\leq7\); BOS & \(N=m+1\), 2 to 8 \\
Dyck 1, Dyck 2 & Fifteen brackets of one or two types; BOS & \(N=16\) \\
Addition & Digits 0 through 9, \texttt{+}, BOS; \(d_a,d_b\in\{1,2\}\) & \(N=d_a+d_b+2\) \\
Addition with hints & Same operands and delimiters; one letter before each digit & \(N=2(d_a+d_b)+2\) \\
\bottomrule
\end{tabular}
\end{table}

Before training, we corrected two issues in the inherited addition generator.
The first tokenized a whole answer as one token instead of separate digits.
The second restricted the second operand to a single digit although the task allowed two.
The corrected generator produces a token for each digit of the answer and allows each operand to have one or two digits.
The corrected natural generator samples digits individually and may produce operands with leading zeros.
The balanced generator samples one digit numbers between 0 and 9 and two digit numbers between 10 and 99 without leading zeros.
Both generators use standard decimal representation for the answers.
Hint letters indicate positions from the left of each field, not shared decimal places across fields of different widths.
The letters restart for each operand and the answer.
For example, \texttt{[<s>,a,9,+,a,9,b,9]} represents \(9+99\) and requires \texttt{[<pad>,a,1,b,0,c,8,<s>]}.
The answer starts at position one after the beginning token (BOS).

Token accuracy measures the proportion of scored target positions that are correctly identified.
Sequence accuracy awards example credits only if all scored positions in the example are correct.
It takes the average over examples.
Token accuracy gives greater weight to longer scored answers, while sequence accuracy assigns one decision to each example.
In the case of Histogram, Double Histogram, sorting and reversal, the scored positions are the ordinary value positions.
In the case of the bracket tasks, all prefix labels are scored.
Boundary targets and batch padding are not included.

In the cases of Frequency Ranking and Addition, target positions left unused in the inherited answers are filled with \texttt{<s>} instead of \texttt{<pad>}, and the historical neural scorer excludes padding but includes this filler.
The Python addition scorer in Section~\ref{sec:evaluation-protocol} instead excludes both padding and filler, and scores the aligned answer digits by position rather than interpreting the prediction as an integer.
The corresponding hints mask retains the letters and digits.
For the frequency program examples, we are concerned only with the required ranks at positions one through \(|U(x)|\) and ignore the remaining filler.

\subsubsection{Training Defaults and Model Selection}
The baseline grid employs between 1 to 4 layers, 2, 4, 8 or 16 heads, and 2, 4 or 8 MLPs per layer.
There are 48 configurations per task, and 5 seeds per configuration across 9 tasks.
Thus, a total of 2160 models were trained.
The seeds were 13877, 27983, 47389, 51229 and 56543.
In \texttt{L\{x\}\_H\{y\}\_M\{z\}}, if \texttt{\{x\}} = 4, then each layer contains \texttt{\{y\}} heads and \texttt{\{z\}} MLPs, and both totals are split equally between categorical and numerical components.

Training hyperparameters are as follows, unless stated otherwise.
Adam optimizer is used with a learning rate of 0.05.
The batch size is 512.
There is no dropout.
Training continues for 250 epochs.
The temperature is decreased geometrically from 3.0 to 0.01.
During training, the model uses continuous relaxations of its discrete choices.
Final hard neural evaluation selects discrete alternatives using \texttt{argmax}.
Bracket tasks use causal attention.
All other tasks, use bidirectional attention.

We are typically given a request for 20,000 examples to train on, but how many we actually use is determined by factors such as duplicate examples and how the data is divided.
For example, depending on whether the natural frequency request is 20,000 or 50,000, we end up with 16,200 or 40,500 examples to train with, respectively.
Equal counts of epochs do not mean we have the same number of updates.

Some results in the main tables are not averages of all the seeds or configurations.
Examples of intermediate results are shown as well as the final results from intervention models.
Grid baselines, intermediate soft peaks and final intervention models are selected separately.
For the frequency baseline, selection of the final result uses the hard token score.
The enriched frequency results shown in Table~\ref{tab:final-results} are soft neural scores, while the later code examples examine the final Python program separately.
Repair checks for histogram and sorting concern hard neural models.
We show results of the system under evaluation, and distinguish soft models, hard neural models and Python programs rather than treating them as interchangeable.

\begin{table}[htbp]
\caption{Models and programs used in the main computational examples and repair checks. The histogram model with 42 exhaustive failures is not the highlighted program from Section~\ref{sec:hist-analysis}.}
\label{tab:app-cases}
\centering\footnotesize
\begin{tabular}{@{}>{\raggedright\arraybackslash}p{0.60\linewidth}cr@{}}
\toprule
Role & \(L/H/M\) & Seed \\
\midrule
Highlighted histogram program & 4/2/2 & 27983 \\
Histogram repair baseline and retrained example & 4/2/2 & 13877 \\
Initial sorting program & 3/16/4 & 27983 \\
Selected sorting model after coverage enrichment & 3/16/4 & 51229 \\
Baseline frequency program & 4/16/8 & 47389 \\
Enriched frequency program & 4/16/8 & 56543 \\
\bottomrule
\end{tabular}
\end{table}

\subsection{Training Interventions}
\label{app:training}
This section provides comparisons for Section~\ref{sec:improving} and describes the training conditions for the outcomes shown.
We first explored greater depth and variations in the training.
As mentioned in Section~\ref{sec:improving}, we adjusted the distribution of the inputs to the model to address observed failure modes.
The comparisons shown here retained combinations of changes that were made together to the training recipe.
Improvements from combinations of changes do not explain the contribution of each individual change.

\subsubsection{Capacity and Training Procedures}
The six layer reference bracket constructions inspired testing of models that are deeper than the baseline grid.
The models with six layers have hyperparameters H16/M8 and are trained for 251 epochs.
Their final hard neural sequence accuracy on 2000 test cases is shown in Table~\ref{tab:app-depth}.
As mentioned in Section~\ref{sec:improving}, the six layer model did not improve on the selected four layer grid result for Dyck 1, and improved only slightly on the selected two layer result for Dyck 2.
Neither of the two tasks was solved.

\begin{table}[htbp]
\caption{Sequence accuracy in percent for greater bracket depth. The grid column is the selected best result, not a seed mean. The five remaining columns are individual six layer runs.}
\label{tab:app-depth}
\centering\footnotesize
\begin{tabular}{@{}lrrrrrr@{}}
\toprule
Task & Grid best & 13877 & 27983 & 47389 & 51229 & 56543 \\
\midrule
Dyck 1 & 98.60 & 97.55 & 97.20 & 90.15 & 96.05 & 94.95 \\
Dyck 2 & 95.75 & 81.90 & 91.70 & 86.50 & 84.40 & 96.50 \\
\bottomrule
\end{tabular}
\end{table}

The mean token accuracy for position coupling~\citep{cho24}, shown in Table~\ref{tab:app-optimization}, did not improve over its comparator in the tested addition case.
Entropy regularization helped the two weakest comparator seeds but hurt the strongest one.
Training for 500 epochs rather than 250 was used for the entropy recipe.
That recipe also requested more data, so the improvement cannot be attributed to regularization alone.
The later variable format endpoints shown in Table~\ref{tab:final-results} are separate from these comparisons.
The addition recipes shown in this section are in the carry balanced format at L3/H16/M8.

Our ATP implementation enabled the proposed changes.
Specifically, we used smooth transition, uncertainty aware attention and position aware numerical attention~\citep{atp}, and modified both the categorical and the numerical embeddings.
A separate cyclic recipe used 3 temperature cycles with an initial temperature of 2 rather than 3.
Isolating the schedule from the data in the comparison was not implemented.
Both recipes performed below their respective comparison baselines.
Additionally increasing the natural data request from 20,000 to 50,000 reduced the mean token score, and all 5 treated seeds scored below all 5 comparison seeds.
The frequency experiments utilized L4/H16/M2, while the main program analyses used L4/H16/M8.

\begin{table}[htbp]
\caption{Final token accuracy by seed, in percent, for the training comparisons discussed in Section~\ref{sec:intervention-method}. Each task group has its own configuration and comparator. Dataset requests do not necessarily equal unique training set sizes.}
\label{tab:app-optimization}
\centering\footnotesize
\setlength{\tabcolsep}{3pt}
\begin{tabular}{@{}lrrrrrr@{}}
\toprule
Recipe & 13877 & 27983 & 47389 & 51229 & 56543 & Mean \\
\midrule
\multicolumn{7}{@{}l}{Addition, \(L/H/M=3/16/8\)} \\
Carry balanced comparator & 90.23 & 92.76 & 89.48 & 94.22 & 91.36 & 91.61 \\
Position coupling & 89.46 & 91.66 & 91.75 & 91.57 & 89.91 & 90.87 \\
Entropy and longer training & 91.79 & 96.98 & 93.32 & 92.71 & 90.34 & 93.03 \\
\midrule
\multicolumn{7}{@{}l}{Frequency ranking, \(L/H/M=4/16/2\), 250 epochs} \\
Natural, request 20,000 & 80.90 & 81.78 & 82.40 & 81.84 & 81.97 & 81.78 \\
ATP implementation & 65.28 & 65.02 & 68.91 & 65.59 & 60.13 & 64.98 \\
Cyclic temperature & 55.82 & 59.33 & 57.43 & 47.63 & 56.70 & 55.38 \\
Natural, request 50,000 & 69.52 & 68.01 & 70.11 & 69.22 & 68.93 & 69.16 \\
\bottomrule
\end{tabular}
\end{table}

\subsubsection{Changing the Training Examples}
Variable format addition uses the length recipe to assign probabilities 0.25, 0.10, 0.10, and 0.55 to operand length pairs \((1,1)\), \((1,2)\), \((2,1)\), and \((2,2)\) respectively.
It adds no zero padding, allows repetition, and does not impose a disjoint partition of the finite input space.
The layer, head and MLP counts in the two length comparisons do not change.
However, they request 50,000 examples each instead of the 20,000 examples.
Also, for addition, the variable dimension changes from 13 to 16.
The selected final recipes change H16 to H24 and modify the training duration (see Table~\ref{tab:app-recipes}).
Thus, the changes to the length probabilities were not the only changes made to the system.

The histogram recipe targets larger counts identified by counterexamples.
At fraction 0.45, the recipe constructs a high count input instead of sampling naturally.
Within that branch, probability 0.6 selects an input whose values are all identical.
Otherwise, one value fills at least half of the input positions, and the remaining positions are sampled randomly.
The input is then randomly shuffled.
The full model is retrained at L4/H2/M2, with variable dimension 8 and 250 epochs.
This supplies more examples for count six and introduces the missing count seven label.

Frequency enrichment increases the request from 20,000 to 100,000, increases variable dimension from 8 to 9 and modifies ordinary input lengths from 1 through 7 to 3 through 7.
It targets a 40\% quota of inputs with at least three distinct values tied at the maximum count.
The enriched distribution is used in both sampled assessment and training.
The two programs illustrated in the main text also have different seeds (see Table~\ref{tab:app-cases}).
Consequently, a comparison of these programs involves combined training recipes and does not isolate the effect of adding ties to training.

The enriched double histogram study uses L4/H16/M8, dimension 9, a batch size of 512, 300 epochs, and a request for 100,000 examples.
The ordinary input length in this case is between 1 and 8.
The sampler aims to achieve maximum counts of 1, 2, 3, 4, and at least 5.
The proportions for the first four categories are 25\%, 30\%, 25\% and 15\% respectively.
The remaining 5\% is aimed at maximum counts of 5 and above.
The sampler uses rejection sampling with a finite fallback.
Temperature decreases geometrically from 3.0 to 0.01 before final \texttt{argmax} evaluation.
As in the case of histogram and frequency enrichment, examples used in training and testing come from the modified distribution.
Independent generation does not enforce disjoint input sets.

\begin{table}[htbp]
\caption{Settings for the addition data recipes. All use dimension 16 and request 50,000 examples. Sequence scores are percentages. The length comparison and selected final recipe are different experiments.}
\label{tab:app-recipes}
\centering\footnotesize
\begin{tabular}{@{}>{\raggedright\arraybackslash}p{0.31\linewidth}crr>{\raggedright\arraybackslash}p{0.25\linewidth}@{}}
\toprule
Recipe & \(L/H/M\) & Seed & Epochs & Sequence result \\
\midrule
Addition length sampling & 4/16/2 & 47389 & 250 & 2.73 to 58.48 \\
Hints length sampling & 4/16/8 & 56543 & 250 & 39.66 to 65.20 \\
Final addition & 4/24/8 & 47389 & 150 & 76.30 \\
Final hints & 4/24/8 & 51229 & 500 & 84.22 \\
\bottomrule
\end{tabular}
\end{table}

\subsubsection{Repair Across Seeds}
\label{app:repair}
To differentiate removal of a few targeted failures from correct behavior across the input space, we examined all legal inputs in the repair domains.
The histogram domain of 335,922 legal inputs with ordinary lengths from 1 to 7, over 6 symbols, is examined.
The sorting domain of 19,530 legal inputs, over 5 symbols and having ordinary lengths from 1 to 6, is also examined.
Both checks make use of hard neural models.
All ordinary output positions are compared, excluding boundary tokens and padding.
Each input is counted as one failure if at least 1 ordinary output position is in error.

The separate histogram baseline with seed 13877 failed on 36/216 inputs where maximum count is 6, and also failed on all 6 inputs where maximum count is 7.
It was successful on 335,700 other inputs.
After enrichment, all 5 seeds had zero failures when the entire domain was considered (Table~\ref{tab:app-repair-seeds}).
Output label 6 was already present, so introducing label 7 alone cannot explain the removal of all 42 failures.

Sorting has a different behavior.
The initial high count recipe with fraction 0.45 corrects, for all tested seeds, all 5 identical inputs of ordinary length 6, but increases overall failures.
Recipes with fractions 0.15 and 0.30 do not help.
A subsequent recipe targeting residual cases of ordinary lengths 5 or 6, with a dominant value appearing 3 or 4 times and other values appearing once, still leaves failures for all seeds, including failures on identical inputs.

The coverage analysis suggested repeated difficult examples displaced distinct training inputs.
We, therefore, kept a deduplicated natural base, and included high count examples and the dominant value failure cases.
At 0.30 this mixture produced 1 clean seed, 51229, while another seed had 68 failures.
The mean failure count across seeds was 15 as compared to 11 for natural training.
Increasing the mixture to 0.50 gave no clean seed, and seed 51229 showed 14 failures.
This was not a repeat under identical conditions, as the recipe was changed.
The success was selected from about 30 exploratory retraining runs, and does not establish reliable sorting repair.

\begin{table}[htbp]
\caption{Exhaustive hard neural failure counts across seeds. Histogram uses 335,922 inputs and sorting uses 19,530. Every sorting condition is checked on the same domain, even when its sampled test distribution differs.}
\label{tab:app-repair-seeds}
\centering\footnotesize
\setlength{\tabcolsep}{4pt}
\begin{tabular}{@{}lrrrrrr@{}}
\toprule
Condition & 13877 & 27983 & 47389 & 51229 & 56543 & Mean \\
\midrule
Histogram enrichment 0.45 & 0 & 0 & 0 & 0 & 0 & 0.0 \\
\midrule
Sorting natural baseline & 13 & 9 & 4 & 9 & 20 & 11.0 \\
Sorting high count 0.45 & 76 & 47 & 33 & 34 & 49 & 47.8 \\
Sorting high count 0.15 & 262 & 85 & 62 & 31 & 67 & 101.4 \\
Sorting high count 0.30 & 579 & 128 & 397 & 29 & 16 & 229.8 \\
Sorting failure cases 0.30 & 13 & 15 & 51 & 22 & 67 & 33.6 \\
Sorting coverage 0.30 & 68 & 1 & 4 & 0 & 2 & 15.0 \\
Sorting coverage 0.50 & 1 & 22 & 5 & 14 & 16 & 11.6 \\
\bottomrule
\end{tabular}
\end{table}

\subsubsection{Variation During Annealing}
\label{app:annealing}
Figure~\ref{fig:double-histogram-annealing} shows results from the enriched 300 epoch double histogram training recipe, not the baseline grid.
In the seed order used above, final sequence scores were 32.01\%, 92.10\%, 77.28\%, 96.90\%, and 100.00\%.
The baseline sequence score for the grid baseline (32.0\% to 97.5\%) varies across the seeds.
Other reported sequence baselines vary from 24.0\% to 39.7\% for addition with hints and from 6.7\% to 19.1\% for frequency rankings; three of the five seeds used for addition attained a score of zero.
These reported ranges help distinguish selected best scores from consistent training outcomes.

For the addition trajectory in Section~\ref{sec:constraints}, seed 27983 attained the best final score of 69.60\% in that study, but was not the seed with the highest soft peak.
Seed 51229 (L4/H24/M8) attained a soft sequence accuracy of 90.50\% at epoch 71 and a final score of 56.12\% at 500 epochs.
The 14.16\% Python answer score was obtained by evaluating the program extracted from the saved epoch 71 checkpoint on 5,000 examples.
It is neither the final model score nor a measurement of the decline during the remaining training epochs.

\subsection{Verification Details}
\label{app:verification}
\subsubsection{Properties and Checking Procedure}
The verification target consists of a particular checkpoint, its extracted program and classifier weights, and a specified input domain.
TPV~\citep{tpv} reconstructs the hard model from its weights and translates the operations into constraints for Z3.
The extracted Python is loaded separately for replay.
For each total input length \(N\), the query fixes the allowed alphabet and boundary structure and asks whether an input violates the task property.

For histogram and double histogram, the property compares the required counts at every ordinary output position.
Sorting must preserve the input multiset and place it in ascending order; reversal must reproduce the exact reversed sequence.
The bracket properties check every prefix label, including persistent invalidity after an illegal closing bracket.
To assess an arithmetic witness, we compare the required digit field after BOS; the hints variant also requires the correct letters.
The historical frequency property checks distinct value coverage and nonincreasing counts, but does not constrain the order of equal count values.
A violation of those necessary conditions is a task failure; satisfying them alone would not establish the full tie rule.

A solver witness is a candidate counterexample until Python replay confirms a wrong required output.
The harness excludes spurious witnesses and searches again, with a limit of twenty refinement attempts.
Reaching that limit, timing out or returning \texttt{unknown} is inconclusive.
An \texttt{unsat} result rules out violations of the encoded property, but applies to the program only if the encoding and property describe its behavior on the claimed domain.
Alternatively, exhaustive execution checks every legal input directly against the task specification.
Both approaches establish only their stated finite coverage; success at one length does not imply success at another.

\subsubsection{Selected Checks and Further Failures}
Table~\ref{tab:app-checks} shows the artifacts chosen for the audit and the results of later repair.
The initial histogram and sorting programs are grid selections and have counterexamples.
Later clean outcomes for histogram and sorting concern retrained hard neural models (refer to Appendix~\ref{app:repair}).
Other examples shown in the table are selected intervention endpoints, rather than another evaluation of an untouched grid cohort.

\begin{table}[htbp]
\caption{Selected checks. All lengths include special tokens as defined in Table~\ref{tab:app-domains}. A failure establishes an incorrect answer for the specified artifact. No failure reported refers only to the stated checking method and domain.}
\label{tab:app-checks}
\centering\footnotesize
\setlength{\tabcolsep}{3pt}
\begin{tabular}{@{}>{\raggedright\arraybackslash}p{0.20\linewidth}cr>{\raggedright\arraybackslash}p{0.48\linewidth}@{}}
\toprule
Task & \(L/H/M\) & Seed & Result \\
\midrule
Initial histogram & 4/2/2 & 27983 & Python failure at \(N=8\). \\
Initial sorting & 3/16/4 & 27983 & Python failure at \(N=8\). \\
Reversal & 4/16/4 & 13877 & Python failures at \(N=5,6\). \\
Dyck 1 & 4/16/8 & 47389 & Replayed failure at trained length \(N=16\). \\
Dyck 2 grid & 2/16/8 & 27983 & Replayed failure at \(N=16\). \\
Dyck 2 depth recipe & 6/16/8 & 56543 & Model test failure at \(N=16\); solver time limit reached. \\
Double histogram & 4/16/8 & 56543 & No failure reported at \(N=2\) through 8: solver through 7, Python enumeration at 8. \\
Frequency ranking & 4/16/8 & 56543 & Replayed failures at \(N=3\) through 7. \\
Addition, 150 epochs & 4/24/8 & 47389 & Python wrong answer at \(N=6\). \\
Hints, 500 epochs & 4/24/8 & 51229 & Python wrong answers at \(N=8\). \\
\midrule
Retrained histogram & 4/2/2 & All five & Zero hard neural failures on 335,922 inputs, \(N=2\) through 8. \\
Retrained sorting & 3/16/4 & 51229 & Zero hard neural failures on 19,530 inputs, \(N=3\) through 8. \\
\bottomrule
\end{tabular}
\end{table}

Table~\ref{tab:app-witnesses} lists a few more examples of failure cases, in addition to histogram and frequency examples described in the main text.
The sorting example in the table shows that the program changes an input value rather than merely placing it at the wrong rank.
The reversal example is a palindrome, but the program changes its middle value.
The example of addition in the table shows the answer digits.
Predicted digits are shown aligned, not concatenated with filler predictions.
These cases are not claimed to be the shortest possible failures.
Rather, they are representative cases of failure of the respective programs.

\begin{table}[htbp]
\caption{Further failing inputs for the initial sorting, reversal and final addition artifacts in Table~\ref{tab:app-checks}. Boundary tokens are omitted. The hints row omits letters to show the wrong arithmetic answer.}
\label{tab:app-witnesses}
\centering\footnotesize
\setlength{\tabcolsep}{4pt}
\begin{tabular}{@{}lclll@{}}
\toprule
Task & \(N\) & Ordinary input & Program answer & Required answer \\
\midrule
Sorting & 8 & \([3,3,3,3,3,3]\) & \([1,3,3,3,3,3]\) & \([3,3,3,3,3,3]\) \\
Reversal & 5 & \([3,1,3]\) & \([3,3,3]\) & \([3,1,3]\) \\
Addition & 6 & \(19+22\) & \(31\) & \(41\) \\
Addition with hints & 8 & \(9+99\) & \(107\) & \(108\) \\
\bottomrule
\end{tabular}
\end{table}

The case studies on bracket failures demonstrate the significance of considering the full prefix specification.
The last prefix of the saved Dyck 1 input after BOS, ``\texttt{(()((())()))())}'', contains a net negative balance, but the program reports \texttt{P} rather than \texttt{F}.
For the saved Dyck 2 witness, ``\verb|({}((()}({){}}}|'', an open parenthesis on the stack does not match the closing brace at position 8.
The program reports \texttt{F} in this case and shifts to \texttt{P} in the next step, even though the prefix is still invalid.

The shortest frequency witness reported in the case studies contained just two ordinary values, and occurs at \(N=3\).
This example reflects a failure of the task specification, but not evidence of failure inside the enriched training support, whose minimum ordinary length is three.
The separate example with seven ordinary values in Section~\ref{sec:most-freq-after} demonstrates a failure inside that support.

\subsubsection{Domains and Verification Cost}
\label{app:length-runtime}
For the example provided for histogram, the full set of ordinary strings of length between one and seven over six symbols requires the model to consider \(335{,}922\) different inputs.
For the example provided for sorting, the full set of ordinary strings of length between one and six over five symbols requires the model to consider \(19{,}530\) inputs.
Checking the exhaustive hard neural repair domains for these examples required evaluation of all legal inputs up to total length eight.

Interestingly, the example provided for sorting contains an extra EOS, explaining the difference in the ordinary lengths of the examples.

The double histogram program's solver checks extend to a total length of seven.
The query at this length took 5,923 seconds, about 100 minutes.
The solver failed to finish within eight hours at total length eight.
The researchers used enumeration in Python to examine all \(279{,}936\) inputs at that length, and reported no counterexamples.
This describes a change to the checking procedure and does not refer to the answer to a solver query at total length eight.
The enumeration covers seven ordinary values and does not extend to the enriched training support's maximum of eight ordinary values.
The researchers noted that the observed cost for this learned program and its encoding does not establish running time laws for all computational tasks and verifications.

\subsection{Supporting Computations}
\label{app:program-analysis}
\subsubsection{The Retrained Histogram Answer}
\label{app:hist-program-records}
As described in the main paper, the retrained histogram program provides seven answers, all equal to seven, to seven copies of the value four.
To determine the classification logic of the program for this input, we manually trace its operations and add the associated classifier coefficients.
The program counts seven at each of the seven ordinary positions and then sums those counts across the selected positions to obtain 49.
Each of these operations are also present in the baseline programs, and therefore, are not new operations learned by the program as a result of the enrichment.
Other components of the model and associated classifier weights are also updated during a complete model retraining.

The score account for the retrained program with seed 13877 at the first of the seven positions is provided in Table~\ref{tab:app-hist-account}.
The first count's score contribution favors class six, but the sum of counts reverses this ordering, and as shown by the final classification scores, class seven is the most preferred class.
The final classification scores for classes one through seven are provided below.

(Class 1: \(-183.554\), Class 2: \(-222.556\), Class 3: \(-89.762\),
Class 4: \(-12.324\), Class 5: \(31.273\), Class 6: \(54.409\), Class 7: \(67.800\))

Thus, among all the competing classes, class seven is the most preferred class.
As shown by the source code provided, operations similar to those shown for the first position occur for the other six positions, with class seven winning and class six the runner up at each.
This manual calculation uses the saved source and decimal classifier coefficients and is separate from the exhaustive hard neural checks.

\begin{table}[htbp]
\caption{Classifier contributions for the retrained histogram at the first ordinary position. The input is seven copies of 4. Values are rounded to three decimals after computing the totals.}
\label{tab:app-hist-account}
\centering\footnotesize
\begin{tabular}{@{}lrr@{}}
\toprule
Contribution & Answer 6 & Answer 7 \\
\midrule
Other contributions combined & $-57.205$ & $-117.785$ \\
First count, value 7 & 8.769 & $-7.019$ \\
Sum of counts, value 49 & 102.845 & 192.605 \\
Total & 54.409 & \textbf{67.800} \\
\bottomrule
\end{tabular}
\end{table}

\subsubsection{Complete Frequency Score Totals}
\label{app:most-freq-analysis}
Most examples in the trace revolve around the answer chosen by each program.
Appendix~\ref{app:records} contains the scores for all available classes at the second position of the output for the inputs \([1, 3, 4, 3, 4, 4, 4]\) (Table~\ref{tab:app-frequency-totals}).
The correct answer in this case is 3.
After combining all the scores, the baseline program chose 4 and the enriched program chose 1.
In the enriched program, the score for 1 is in fact greater than the score for 4 by about 0.00472.
Therefore, answering before rounding is pertinent.
These scores are obtained from saved traces of the program and not from a new evaluation.

\begin{table}[htbp]
\caption{All classifier totals at the second output position of the frequency example. The main tables separate a selected count contribution from the remaining scores; this table checks the winner against all classes. BOS is not an output class in the enriched program.}
\label{tab:app-frequency-totals}
\centering\footnotesize
\begin{tabular}{@{}lrr@{}}
\toprule
Answer class & Baseline total & Enriched total \\
\midrule
0 & $-5.15765$ & 6.73346 \\
1 & 10.88742 & \textbf{13.77100} \\
2 & 3.82967 & $-11.90848$ \\
3 & 7.30950 & 9.03644 \\
4 & \textbf{14.35276} & 13.76628 \\
5 & $-8.65461$ & $-21.47170$ \\
BOS & $-22.27695$ & Not available \\
\bottomrule
\end{tabular}
\end{table}

Each of the frequency programs is made up of 96 learned components, made up of 16 attention heads and 8 MLPs in each of the 4 layers.
The baseline is made up of 2145 lines of code, and the enriched program is made up of 2022 lines of code.
Blank lines, comments and the example are included in these counts, but the classifier CSV is not.
These counts describe the two inspected programs.
The components of the programs are not always easy to understand.
The score justifications show reasons as to why answers are wrong, but do not provide the full justification of the ranking.
It is important to note that separating a score term does not remove the corresponding feature from later computations.
This is not an ablation of every path through that feature.

\subsubsection{Reference Construction Costs}
\label{app:constructions}
Reference programs help distinguish failure to learn a solution from inability to represent one.
Table~\ref{tab:app-constructions} summarizes the seven implemented constructions.
The counts describe allocated heads per layer, including heads unused by the intended computation in some layers.
Unlike the training grid, these builders need not divide categorical and numerical heads equally.
MLP width, input arity and variable dimension must also accommodate the chosen input bound; matching only layers and total heads is insufficient.

\begin{table}[htbp]
\caption{Sufficient structural costs of the implemented reference programs. \(H_c\) and \(H_n\) are categorical and numerical heads per layer. \(V\) is the ordinary value vocabulary size and \(n\) is the operand digit width. These are construction costs, not minima.}
\label{tab:app-constructions}
\centering\footnotesize
\begin{tabular}{@{}lrrr>{\raggedright\arraybackslash}p{0.43\linewidth}@{}}
\toprule
Task & Layers & \(H_c\) & \(H_n\) & Computation \\
\midrule
Histogram & 1 & 0 & 1 & Equality count and count label lookup \\
Reversal & 2 & 1 & 0 & Locate end, compute mirror position, copy token \\
Sorting & 3 & 0 & \(V\) & Value counts, cumulative counts and lookup decoding \\
Dyck 1 & 6 & 1 & 2 & Causal balance and bracket matching \\
Dyck 2 & 6 & 1 & 4 & Same construction with two bracket types \\
Addition & \(n+2\) & 1 & 0 & Carry propagation with fixed width output \\
Frequency ranking & 2 & \(V\) & \(V\) & Counts and first positions, then finite lookup \\
\bottomrule
\end{tabular}
\end{table}

The addition construction gathers aligned operand digits, computes their local carry states, passes carry information through \(n\) layers and combines it with the base digit sums.
These states distinguish generating, blocking and propagating a carry~\citep{quirke24}.
This gives \(n+2\) layers for the particular implementation.
Its operands are padded to exactly \(n\) digits and the answer has \(n+1\) digits, including a leading zero when needed.
The input length is \(2n+2\), including BOS and the plus sign.
It therefore demonstrates an arithmetic construction, not identical formatting behavior to the variable width learned task.

The frequency construction counts each value and gathers its first occurrence position.
A lookup table reads those statistics together with the requested output position and returns the corresponding ranked value.
The reported table growth is roughly \(N^V\) for sequence bound \(N\) and vocabulary size \(V\).
Constant depth does not make this a constant size model as the domain grows.
Likewise, the histogram decoder must support the required count labels, and the bracket lookups must represent balances up to the chosen bound.
These examples explain costs of the constructions we used without proving that more efficient constructions are impossible.

\subsection{Computational Resources}
\label{app:compute}
The baseline grid comprised 2,160 training runs: 48 configurations for nine tasks with five seeds each.
The histogram and sorting repair study used 35 further training runs, covering one histogram recipe and six sorting recipes with five seeds each.
Training used CUDA GPUs, with NVIDIA L4, A100 40 GB and T4 devices documented across the intervention studies.
Multiple training workers shared devices.
Table~\ref{tab:app-compute} reports approximate GPU hours and recorded study durations, not a project total.

\begin{table}[htbp]
\caption{Training runs and approximate compute. The first two rows are extrapolated GPU hours under the assumptions below; remaining entries are elapsed execution windows. The estimated ranges describe alternative scenarios, not measured bounds or confidence intervals.}
\label{tab:app-compute}
\centering\footnotesize
\begin{tabular}{@{}>{\raggedright\arraybackslash}p{0.37\linewidth}r>{\raggedright\arraybackslash}p{0.23\linewidth}>{\raggedright\arraybackslash}p{0.22\linewidth}@{}}
\toprule
Study & Runs & Hardware & Time \\
\midrule
Baseline grid & 2,160 & GPU workers & About 260 to 415 GPU hours \\
Histogram and sorting repair & 35 & T4, five workers & About 3 to 8 GPU hours \\
Enriched frequency and double histogram & 10 & Shared GPU VM & 11.19 hours \\
Addition length sampling & 10 & A100 40 GB & About 2.5 hours \\
Addition and hints, 150 epochs & 10 & A100, ten workers & About 45 minutes \\
Bracket models with six layers & 10 & A100 40 GB & About 2.8 hours \\
\bottomrule
\end{tabular}
\end{table}

The grid estimate scales the enriched run timings to 20,000 nominal examples and 250 epochs, assuming comparable hardware and 16 workers per GPU rather than across the fleet.
Its endpoints assume either unchanged elapsed time per run or unchanged device throughput as worker sharing increases.
The repair estimate uses T4 sorting timings as a proxy for all 35 runs, with five workers and alternative assumptions about reduced contention.
These estimates exclude idle allocation and failed attempts.

The ten enriched frequency and double histogram runs overlap: their durations sum to 105.33 hours, but their combined execution window is 11.19 hours.
Assuming one shared L4 throughout gives an estimate of 11.19 GPU hours for these ten runs, not an allocation measurement or a project total.
Timings can include data preparation, evaluation and export.
Complete project timing and allocation records are unavailable.

CPU verification is separate from GPU training.
The double histogram query at \(N=7\) took 5,923 seconds; the \(N=8\) attempt exceeded eight hours before Python enumeration.
Together, these attempts took at least about 9.65 elapsed hours, excluding enumeration, replay and other queries.
CPU core hours are unknown because active core counts were not recorded.

\end{document}